\documentclass[10pt,twocolumn,journal]{IEEEtran}

\usepackage{epsfig}
\usepackage{amsmath}
\usepackage{amssymb}
\usepackage{color}
\usepackage{url}
\usepackage{multirow}
\usepackage{algorithm}
\usepackage{algorithmic}
\usepackage{graphicx}
\usepackage{subfigure}
\usepackage{enumerate}
\usepackage{lineno}

\newlength{\figurewidth}
\newlength{\smallfigurewidth}

\begin{document}

\title
{
SAR Target Recognition Using the Multi-aspect-aware Bidirectional LSTM Recurrent Neural Networks
}

\author{%
Fan Zhang,~\IEEEmembership{Senior Member,~IEEE},
Chen Hu, Qiang Yin,~\IEEEmembership{Member,~IEEE}, Wei Li,~\IEEEmembership{Senior Member,~IEEE}, Hengchao Li,~\IEEEmembership{Senior Member,~IEEE}, and Wen Hong,~\IEEEmembership{Senior Member,~IEEE}
\thanks{This work was supported in part by the National Natural Science
Foundation of China under Grant No.~$61501018$, Grant No.~$61571033$, by the Beijing Natural Science Foundation under Grant No.~$4164093$, and by the National Major Research High Performance Computing Program of China under Grant No.~2016YFB0200300.}
\thanks{F. Zhang, Q. Yin and W. Li are with the College of Information Science \& Technology, Beijing University of Chemical Technology, Beijing $100029$, China. (Corresponding author, e-mail:$yinq@mail.buct.edu.cn$).}
\thanks{C. Hu is with the National Research Center for High-Performance Computing Engineering Technology, Sugon Information Industry Co., Ltd, Beijing $100193$, China.}
\thanks{H.C. Li is with the Sichuan Provincial Key Laboratory of Information Coding
and Transmission, Southwest Jiaotong University, Chengdu, $610031$, China.}

\thanks{W. Hong is with the Institute of Electronics, Chinese Academy of Sciences, Beijing, $100190$, China.}

}

\maketitle
\thispagestyle{empty}
\pagestyle{empty}


\begin{abstract}
The outstanding pattern recognition performance of deep learning brings new vitality to the synthetic aperture radar (SAR) automatic target recognition (ATR). However, there is a limitation in current deep learning based ATR solution that each learning process only handle one SAR image, namely learning the static scattering information, while missing the space-varying information. It is obvious that multi-aspect joint recognition introduced space-varying scattering information should improve the classification accuracy and robustness. In this paper, a novel multi-aspect-aware method is proposed to achieve this idea through the bidirectional Long Short-Term Memory (LSTM) recurrent neural networks based space-varying scattering information learning. Specifically, we first select different aspect images to generate the multi-aspect space-varying image sequences. Then, the Gabor filter and three-patch local binary pattern (TPLBP) are progressively implemented to extract a comprehensive spatial features, followed by dimensionality reduction with the Multi-layer Perceptron (MLP) network. Finally, we design a bidirectional LSTM recurrent neural network to learn the multi-aspect features with further integrating the softmax classifier to achieve target recognition. Experimental results demonstrate that the proposed method can achieve $99.9\%$ accuracy for 10-class recognition. Besides, its anti-noise and anti-confusion performance are also better than the conventional deep learning based methods.
\end{abstract}



\section{Introduction}
\label{sec:intro}
%
%

Due to the imaging characteristics of day-and-night and weather-independent, synthetic aperture radar (SAR) has been widely used for Earth remote sensing for more than $30$ years, and has come to play a significant role in geographical survey, climate change research, environment and Earth system monitoring, multi-dimensional mapping and other applications \cite{RMor,bobo-1}. With the evolving of SAR technologies, massive SAR images with abundant characteristics (e.g., high resolution, multi-aspect, multi-dimension, multi-polarization) have been provided for further applications in the Earth Observation. Different from the corresponding optical counterparts in many aspects, such as speckle noise, backscattering oriented pixel intensity representation, geometric distortions, high sensitivity to the target position and so on \cite{ATRREVIEW}, SAR image is relatively more difficult to interpret. To bridge the SAR systems and their applications, SAR image automatic interpretation, especially automatic target recognition (ATR), has become an important research topic in surveillance, military tasks, etc., and has been studied continuously for more than $20$ years \cite{MSTAR}.

In current high resolution SAR imaging era, the target feature is more of significance such that the ATR research is more focused on the target type identification. The SAR target recognition mainly consists of three steps: pre-processing, feature extraction, and classification. The pre-processing step includes filtering \cite{PROCLee} and segmentation \cite{PROCLevel}, which is employed to provide pure target region in SAR image. The feature extraction aims to reduce redundant information of target image while keeps accurate target representation. On the basis of the former two steps, the classification step tends to get the exact category information by the classifier. The feature selection and design of classifier are the most important parts of SAR ATR algorithms. In terms of feature extraction, the widely used features can be described as the static features, which are selected from an independent image, like geometric feature, scattering feature, polarization feature, transform domain feature and so on \cite{GEOF,SCAF,POLF,TRANSF}. As for the classification, there are mainly four kinds of methodologies, including template matching, model-based method, neural networks, and machine learning \cite{GEOF,ATRSVM,ATRBOOST,ATRMODEL,ATRANN}. The template matching method is practical, but is more dependent on the construction of template library. If the SAR sensor changes, the accuracy will drop rapidly. To avoid the construction of template library, the model-based method employs the high-fidelity model to represent the target feature instead of image template, but its adaptability is still limited. The traditional neural network has the disadvantages of poor learning performance, big training data requirement. Compared with these methods, the machine learning based ATR proves to be stable, efficient, and accurate, for example, support vector machine (SVM) \cite{ATRSVM}. Although SVM seeks to separate classes by learning an optimal decision hyperplane that best separates training samples in a kernel-induced high-dimensional feature space, there are some issues hindering its applicability, e.g., the selection of kernel function parameters, classification speed.

Despite numerous ATR research over the past thirty years, very few ATR algorithms have been applied into practical applications. One of the most important reasons is the poor false-alarm performance, which is related with the feature representation and classifier. Intrinsically, SAR image feature is a space-varying scattering feature, which changes dramatically with the variations of aspect and depression angle. Usually, the depression angle is known in advance, the aspect angle information is highly demanded for ATR application. Correspondingly, some aspect estimation algorithms are proposed to handle this issue \cite{AspectEst-1,AspectEst-2}. Even so, the classifiers trained with a specific aspect interval may not perform well if the test image falls out of this interval. To alleviate this problem, the complexity of classification has to be increased by multiple classifier methods, such as ensemble classifiers on different aspect angle, majority-voting strategy and so on \cite{MVATR}. So this kind of multi-aspect joint recognition seems to be a solid method to decrease the false-alarm rate, on the other hand, it may put forward higher requirements for the data acquisition. In fact, we can often obtain the multi-aspect SAR images in practice, i.e., multiple airborne/UAV SAR joint observation in different aspect angles, single SAR observations along a curvilinear or circular orbit. Correspondingly, the collected images can cover multiple aspect angles or even all-aspect angles, bring the comprehensive representation for target scattering signature, and provide one possibility to improve the recognition performance.



With the richness of multi-aspect signature, the classification technology is also further developed with the concept of deep learning. In recent years, the emerging deep learning methods demonstrate their excellent target recognition capability, and have been widely applied in signal processing, image processing, and remote sensing \cite{DLHinton,DLHinton1,DLSig,DLHyp1,DLHyp2,DLFZ1,DLFZ2}. Also, the deep learning based SAR image processing has become a hot topic and outperformed the traditional approaches. For a deep neural network, a large amount of data sets are required to train millions of network weights. Compared with limited SAR ATR data, SAR image data for classification can fairly well meet the requirement and has been effectively classified by deep convolutional autoencoder (CAE) \cite{DLSARIC1}, deep belief network (DBN) \cite{DLSARIC2}, restricted Boltzmann machine (RBM) \cite{DLSARIC3}, convolutional neural networks (CNN) \cite{DLSARIC4} and so on. As for the SAR ATR application, shallow CAE and CNN are employed to conduct MSTAR and TerraSAR-X data recognition with sparsely connected convolution architectures and fine-tuning strategies, and achieve high recognition accuracy over $98\%$ in type identification \cite{DLSARJYQ,DLSARDLR,DLSARXHP}. However, for configuration recognition and confuser rejection, the recognition accuracy is reduced to $87\%$. Therefore, the deep learning  based SAR ATR still be worthy of further studying.

The key perspective of deep learning is that these layers of features are not designed by human engineers: they are learned from data using a general-purpose learning procedure \cite{DLHintonNature}. Different from natural image in data and feature representation, SAR image brings two problems for deep learning based ATR solution. Firstly, SAR imagery is essentially a coherent image indicating the coherent backscatter. Secondly, the learned multi-level features are still limited to the static features, which are extracted from an independent image. Therefore, there is a most straightforward thought to decrease the false-alarm rate by using the space-varying scattering feature from the multi-aspect image sequence instead of the static feature from a single image. The multi-aspect even all-aspect images may offer such opportunity to construct the image sequence, which are employed to extract space-varying backscattering feature. This aspect-dependence property in SAR image is especially notable due to the physical composition of the target. Therefore, the sequential information that extracted from the multi-aspect images at a single target, is capable of offering the potential to substantially improve identification performance \cite{MAHMM1}. In the multi-aspect information based ATR, the hidden Markov models (HMMs) based methods are the mainstream solutions to model such sequential data \cite{MAHMM1,MAHMM2,MAHMM3}. Due to the fact that it is difficult to find a relationship between the backscattering and the HMM state, its application has some limitations, e.g., the HMM modeling is basically derived from the practical experience and is not so solid for other scenarios.

In this work, a novel ATR framework based on multi-aspect-aware bidirectional Long Short-Term Memory (LSTM) recurrent neural networks (MA-BLSTM) is proposed to improve the recognition performance by exploiting the sequential features of multi-aspect views on a single target. LSTM recurrent neural networks were originally introduced for sequence learning \cite{LSTM1}. After strengthened by bidirectional design, bidirectional LSTM has complete, sequential information about all points before and after it \cite{BLSTM-1}. For SAR ATR, these networks including recurrently connected cells learn the dependent features among multiple aspect images, then transfer the probabilistic inference to the next and previous aspect image units. Recent works have indicated that LSTM outperforms hidden Markov models (HMMs) in modeling the stochastic sequences \cite{LSTM2,LSTM2-1}, which was dominated by HMMs in the early $2000s$. It is predictable that LSTM is suitable for multi-aspect feature based SAR target recognition. For each aspect image, the global Gabor features and the three-patch local binary pattern (TPLBP) features are combined in different orientations to extract more comprehensive spatial information, which is employed for multiple aspect images and further constructs the multi-aspect features. Furthermore, a fully-connected Multi-layer Perceptron (MLP) network for feature dimensionality reduction is integrated with the bidirectional LSTM recurrent neural networks to realize a highly efficient SAR ATR. Compared to the state-of-the-art deep learning based ATR methods \cite{DLSARJYQ,DLSARDLR,DLSARXHP}, we make the following contributions.
\begin{itemize}
\item We present an idea of deep learning based ATR method, that is, considering multi-aspect image sequence based joint recognition instead of single aspect static feature based isolated recognition, which can be applied to multi-polarization, multi-static, multi-channel, multi-band and so on.
\item We introduce the bidirectional LSTM recurrent neural network to memorize the context information for multi-aspect sequence data for learning the space-varying scattering feature. The static multi-orientation spatial information with gray-scale and rotation invariant characteristics are extracted from each image as the single aspect features, and further construct the pure multi-aspect features by concatenating all the static features in the sequence.
\item We propose a novel ATR framework including single aspect features extraction by combining Gabor and TPLBP features, supervised feature dimensionality reduction with the MLP, multi-aspect features learning with the bidirectional LSTM, and softmax classifier, which tends to be discriminative and solid, and further improve the recognition accuracy as high as $99.9\%$.


\end{itemize}

The rest of this paper is organized as follows. Section \ref{sec2} specifically introduces the proposed MA-BLSTM ATR framework. Then, the experimental results and analysis are presented in Section \ref{sec3}. Finally, conclusions are drawn in Section \ref{sec4}.

\section{The Proposed ATR Approach}
\label{sec2}

In SAR image formation, the multi-aspect signatures are integrated to form a single image, thereby losing some of the explicit aspect dependence \cite{MAHMM2}. Correspondingly, most mainstream ATR approaches only perform target classification based on a single view of the target. Due to the received signals from different targets may be similar at certain aspects, the space-variant backscattering feature bring troubles for reliable target recognition. The current ATR methods, e.g., the machine learning or deep learning, mainly focus on the independent backscattering feature rather than continuous space-variant backscattering features. On the other hand, to meet the practical demand for ATR in limited training data and confusing environment, the space-variant scattering features, namely the multi-aspect features, may be more essential and solid than separate backscattering feature. Therefore, the theoretical and practical demands motivate us to exploit the multi-aspect sequence features to further improve the recognition performance and adaptability.

\begin{figure*}[!tp]
\centering
\includegraphics{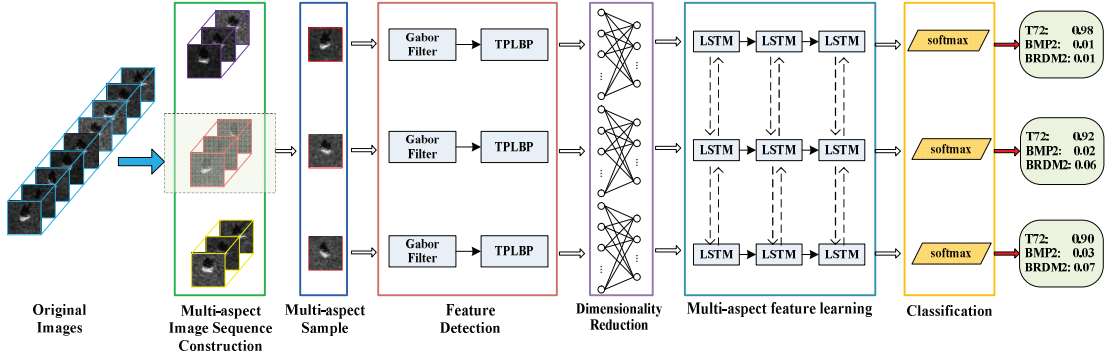}
\caption{A flowchart of the proposed ATR framework based on multi-aspect-aware bidirectional LSTM. }\label{fig1}
\end{figure*}

The proposed target recognition framework is illustrated in Fig.\ref{fig1}, including five processing steps: constructing the multi-aspect image sequence, extracting the separate backscattering feature using Gabor filter and TPLBP approach, reducing the feature dimensionality with a fully-connected MLP network, learning the multi-aspect features with the LSTM recurrent neural networks, and determining the category for each target sample via the softmax classifier.

\subsection{Multi-aspect Image Sequence Construction}
\begin{figure*}[!tp]
\centering
\includegraphics[width=140mm]{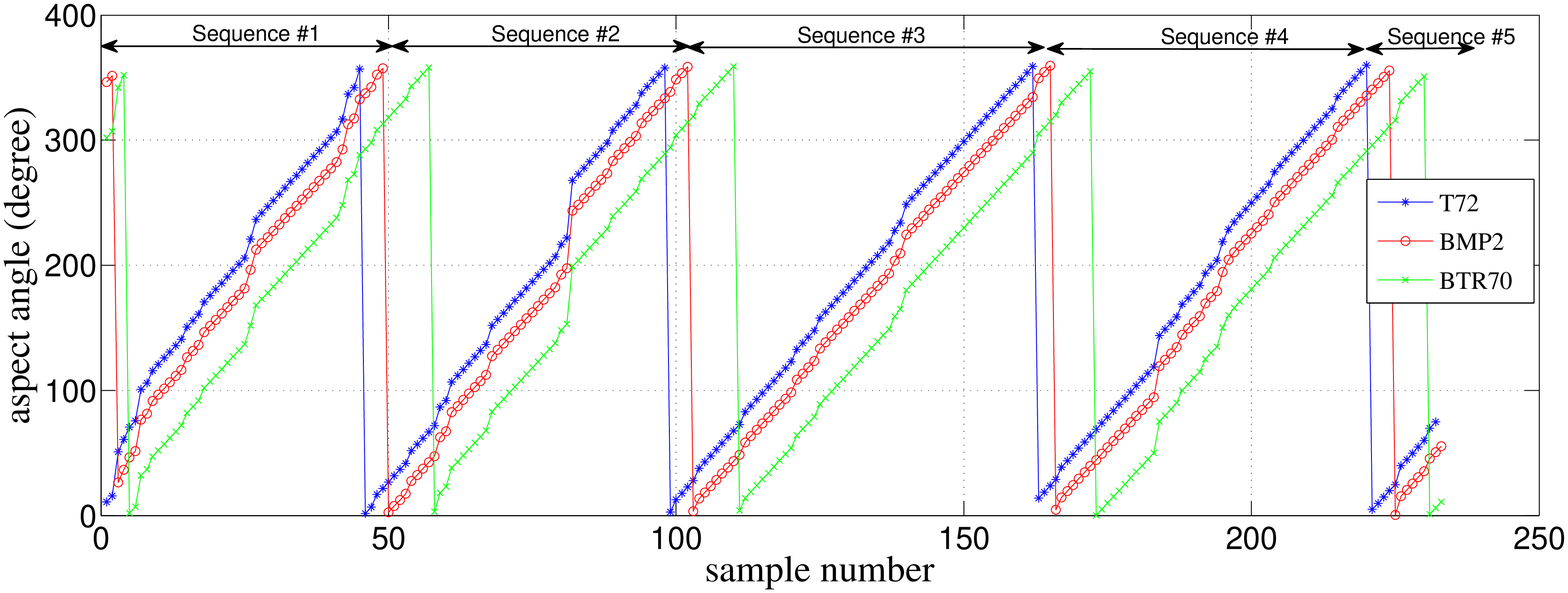}
\caption{Sketch map of Multi-aspect image sequence construction. }\label{figconstruction}
\end{figure*}

In order to model the dynamic multi-aspect features of SAR targets, the corresponding image sequences with continuously varying aspect angles should be selected and constructed from the original data set. This kind of image sequence that can provide the multiple aspect context information of single target is also required as the input sample by the proposed deep bidirectional LSTM networks. As for the popular deep neural networks, e.g., convolutional neural networks (CNN) and stacked autoencoders (SAE), the required input is usually one independent image. Therefore, it is difficult for them to learn the space-varying signature of the targets of interest. In addition to cover the sufficient range of aspect angle, it is better to select the target images with continuous varying aspect to construct a multi-aspect sequence sample. For example, Moving and Stationary Target Acquisition and Recognition (MSTAR) Program provides the data sets with continuous varying aspect, and can be counted for more than four completed circle. Then, $4$ completed circle images are constructed for sequence sample $No.1$ to $No.4$, and the rest images are organized as sequence $No.5$, as shown in Fig. \ref{figconstruction}.

\subsection{Multi-orientation Spatial Feature Extraction}
Magnitude SAR imagery is orientation-sensitive, gray-scale variant; moreover, it has the complex geometric distortions and a fixed scale irrespective of range to the sensors. These special characteristics are originated from the space-varying scattering, various imaging condition, and SAR imaging mechanism, and are often the important considerations for feature selection. Gabor features are capable of representing the orientation and scale of physical structures, while TPLBP features can capture the local spatial information with gray-scale and rotation invariant characteristics. Thus, it is promising to combine Gabor and TPLBP to achieve the relative completed and intrinsic feature representation for each single aspect image.

The Gabor filter can be viewed as an orientation dependent bandpass filter, which is orientation-sensitive and rotation-variant. In order to convert rotation variant features to rotation invariant features, a circularly symmetric Gabor filter is commonly employed, in which all directions for each pass band are considered. The magnitudes of each Gabor-filtered image reflect signal power in the corresponding filter pass band and are used as Gabor features \cite{LBP_LW}, which are the results of convolving the image with the Gabor filters of different orientations. The 2D Gabor filter is a Gaussian kernel function of a sinusoidal plane wave modulation, and are defined as follows \cite{zhoujiarui2012facerecognitionusing}:

\begin{equation}\label{equation1}
\begin{split}
g(x,y,\lambda ,\theta ,\psi ,\sigma ,\gamma ) = \exp \left( { - \frac{{x{'^2} + {\gamma ^2}y{'^2}}}{{2{\sigma ^2}}}} \right)\\
 \quad \quad \quad \times \exp \left( {i\left( {2\pi \frac{{x'}}{\lambda } + \psi } \right)} \right)
\end{split}
\end{equation}
with
\begin{equation}\label{equation2}
\left\{ \begin{array}{l}
x' = x\cos \theta  + y\sin \theta \\
y' =  - x\sin \theta  + y\cos \theta
\end{array} \right.
\end{equation}
where $\lambda$ is the sinusoidal plane wave length, $\theta$ is the orientation of Gabor kernel function, $\psi$ is the phase shift, $\sigma$ is the standard deviation of Gaussian envelope, $\gamma$ is the ratio of $x'$ and $y'$ direction, $(x, y)$ is the image pixel coordinates. Assuming $6$ orientations are employed, one image will be convoluted with $6$ Gabor filters, which will calculate $6$ magnitude values for each pixel and result in $6$ Gabor feature images.

\begin{figure}[!tp]
\centering
\includegraphics[width=40mm]{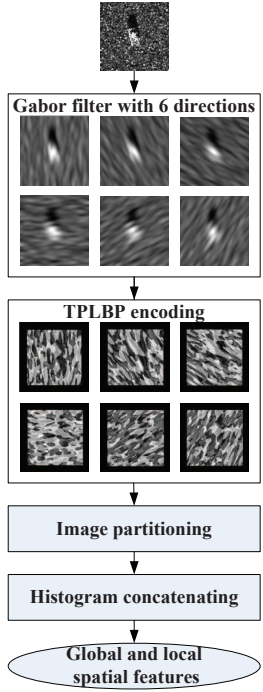}
\caption{The multi-orientation spatial feature extraction using Gabor and TPLBP. }\label{fig2}
\end{figure}

For SAR imagery, Gabor features capture the global spatial information corresponding to selective orientation. These global features are gray-scale variant, and have limited detail spatial information, so they can be further encoded. In order to enhance the local spatial information, we encode the Gabor features with TPLBP operator \cite{TPLBP}, which is derived from the local binary pattern (LBP) method \cite{LBP}. The LBP method uses short binary strings to encode features of the local spatial information around each pixel, and is a gray-scale and rotation invariant feature operator. As an extended version of LBP, the TPLBP code is generated by comparing the value of three patches to produce a single bit value assigned to each pixel, and can better characterize the local structural information. The TPLBP code of each image pixel is calculated as the following
\begin{equation}\label{equation3}
{\rm{TPLB}}{{\rm{P}}_{r,S,w,\alpha }}(p) = \sum\limits_i^S{f(d({{\rm{C}}_i} - {{\rm{C}}_p}) - d({{\rm{C}}_i}_{ + \alpha \bmod S},{{\rm{C}}_p})){2^i}}
\end{equation}
where ${\rm{C}}_i$ and ${{\rm{C}}_i}_{ + \alpha \bmod S}$ are two patches along the ring, ${\rm{C}}_p$ is the central patch, $d(.)$ is the Euclidean distance, $w$ is the patch size, $S$ is the number of additional patches distributed uniformly in a ring of radius $r$ around central patch $p$, $\alpha$ is the interval of the patches to be compared, and $f$ is defined as:

\begin{equation}\label{equation4}
f(x) = \left\{ \begin{array}{l}
1{\rm{,\quad        }}x \ge \tau \\
0{\rm{,\quad        }}x < \tau
\end{array} \right.
\end{equation}

After the TPLBP encoding, the image is divided into a grid of non-overlapping blocks and a histogram measuring the frequency of each binary code is computed for each block \cite{TPLBP}. Then, these histograms are normalized and concatenated to a vector, which is the local spatial feature of a Gabor feature image. In this way, we encode the local spatial information among multi-orientation Gabor feature images and concatenate to a single vector, which represents the completed global Gabor and local TPLBP spatial features of a SAR image, as shown in Fig. \ref{fig2}.

\subsection{Feature Reduction with Shallow Neural Networks}
Once the Gabor and TPLBP based features are obtained, we proceed to reduce the dimensionality of feature space to pursue an efficient feature learning and avoid the curse of dimensionality. The aim of dimensionality reduction is to minimize information loss while maximizing reduction in dimensionality \cite{lam1999feature}. High-dimensional data can be converted to low-dimensional data by training a multi-layer neural network with a small central layer to reconstruct high-dimensional input vectors \cite{DLHinton1}. Therefore, a MLP neural network is introduced to reduce the dimensionality of the extracted multi-orientation spatial features.

\begin{figure}[!bp]
\centering
\includegraphics[width=80mm]{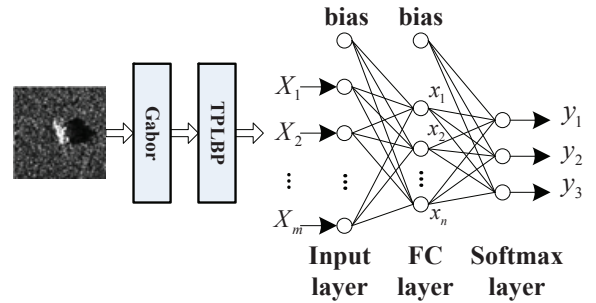}
\caption{The full-connected MLP structure for dimensionality reduction. }\label{fig3}
\end{figure}

The employed MLP neural network is a single hidden layer fully-connected feed-forward network, which consists of an input layer, a hidden layer and an output layer in training phase, and keeps the first two layers in feature reduction phase, as shown in Fig. \ref{fig3}. In the input layer, the number of input units is equal to the dimensionality of the underlying feature extracted from a SAR image. In the hidden layer, the number of neurons is equal to the dimensionality of the input layer of LSTM neural networks. To train the weights between input and hidden layer, we add an output layer, where the number of output units is equal to the number of target categories. The MLP neural network model can be expressed as the following,

\begin{equation}\label{equation5}
f({y_i}|X) = softmax[{b^{(2)}} + {W^{(2)}}(R({b^{(1)}} + {W^{(1)}}X))]
\end{equation}
where $X$ is the input feature vector of a SAR image, $y_i$ indicates class $i$, $f(\cdot)$ is the probability of $X$ belongs to class $i$, $softmax$ represents the softmax function, $b^{(k)}$ is the bias of the $k^{th}$ layer, $W^{(k)}$ is the wights between the $k^{th}$ and $(k+1)^{th}$ layer, and $R$ indicates the Rectified Linear Unit (ReLU) activation function.

During the training stage, the fully connected weights, namely $W^{(1)}$ and $W^{(2)}$, are initialized with random values and adjusted by backpropagating the error between the actual category and its decision. The training process is repeated until the learning error falls below a moderate tolerance lever, such as $10\%$. Then, the reduced features $x$ can be achieved by the feed-forward MLP network with an input layer and a hidden layer,
\begin{equation}\label{equation6}
x(W^{(1)},X) = R({b^{(1)}} + {W^{(1)}}X)
\end{equation}

\subsection{Multi-aspect Features Learning with LSTM}
\begin{figure*}[!t]
\centering
\includegraphics[width=140mm]{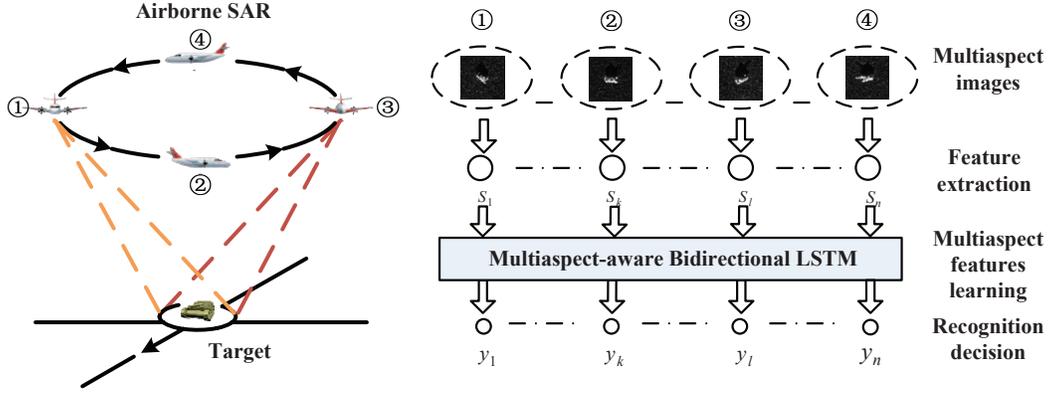}
\caption{Illustration of multi-aspect joint recognition. }\label{fig9}
\end{figure*}

Compared with the separate image feature selection of the main stream ATR methods, the multi-aspect feature selection from several images can obviously increase the recognition rate. The existing researches have shown that the result from multi-aspect classfication is $17\%$ higher in recognition rate than that from single aspect classification \cite{mitchell1999robust}. In order to model the context information among different aspect images, we apply the bidirectional LSTM (BLSTM) networks, which consist of two recurrent network layers in each hidden layer, whereas the first one processes the sequence forwards and the other processes it backwards, as shown in Fig. \ref{fig1}. Since both networks are connected to the same output layer, BLSTM is capable of accessing the entire information about past and future data points in the multi-aspect sequence. Compared with the employed static features in the traditional ATR methods, the BLSTM can easily memorize the context information for multi-aspect sequence data to achieve the space-varying scattering features learning, as shown in Fig. \ref{fig9}. Theoretically, it can achieve higher recognition accuracy.

The proposed multi-aspect-aware bidirectional LSTM networks are divided into three main layers: input layer, hidden layer, and output layer. The input data vectors ($1024\times 1$) are reduced features of the multi-aspect sequence images, which are from the output of MLP networks. The size of input layer is set to be $1024$. The hidden layer includes three BLSTM layers, whose size respectively are $512$, $256$ and $128$. The output layer is the $softmax$ layer, whose size is equal to the number of target classes, and yields the recognition accuracy. In general, the merits of BLSTM are reflected in the LSTM unit and the recurrent property, which will be discussed in the following subsections.

\subsubsection{LSTM memory block}
The basic LSTM unit is composed of three gates (input ${i_n}$, forget${f_n}$, and output ${o_n}$), a single cell ${c_n}$, block input ${I_n}$, an output activation function ${O_n}$, three peephole connections (${p_i}$, ${p_f}$, ${p_o}$) among cell and three gates. The input and output gates scale the input and output of the cell, namely control whether the input signals have an effect on the cell, and whether the cell can impact other neurons. The forget gate scales the internal state of the cell, namely controls whether the state should be remembered or not. The peephole connections scale the state of three gates with the cell state. Finally, the output of the block is recurrently connected back to the block input and all of the gates. These additional gates and peephole connections enable LSTM to model extremely complex and long-term dynamic features, and to overcome the vanishing gradients problems.

\begin{figure}[!tp]
\centering
\includegraphics[width=70mm]{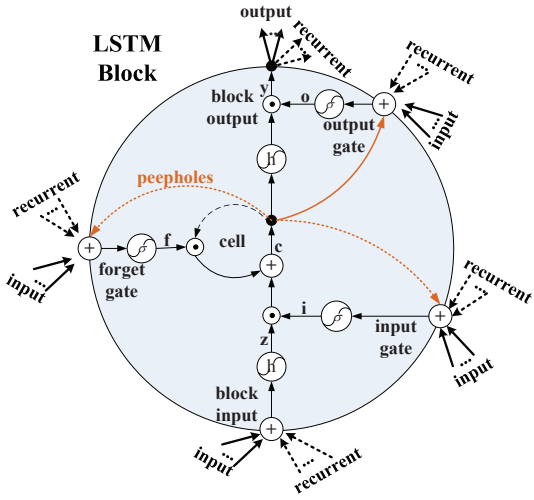}
\caption{Detailed diagram of the LSTM block \cite{greff2015lstm}. }\label{fig4}
\end{figure}

The forward pass of LSTM layer, that is the multi-aspect feature modeling, can be expressed as following formulae, which respectively represent the $block~input$, $input~gate$, $forget~gate$, $memory~cells$, $output~gate$, and $block~output$.
\begin{equation}\label{equation 7}
\begin{split}
& {I_n} = h({W_{I}}{x_n} + {R_{I}}{O_{n - 1}} + {b_I})\\
& {i_n} = \sigma ({W_{i}}{x_n} + {R_{i}}{O_{n - 1}} + {p_i} \odot {c_{n - 1}} + {b_i})\\
& {f_n} = \sigma ({W_{f}}{x_n} + {R_{f}}{O_{n - 1}} + {p_f} \odot {c_{n - 1}} + {b_f})\\
& {c_n} = {i_n} \odot {I_n} + {f_n} \odot {c_{n - 1}}\\
& {o_n} = \sigma ({W_{o}}{x_n} + {R_{o}}{y_{n - 1}} + {p_o} \odot {c_n} + {b_o})\\
& {O_n} = {o_n} \odot h({c_n})
\end{split}
\end{equation}
where $n$ is the order number of the multi-aspect data sequence, $x_n$ is the input feature at $n$-$th$ aspect angle; $W$ are the weight matrices, $R$ are the recurrent weight matrices, $b$ are the bias vectors, $p$ are the peephole weight vectors, the subscripts $\{I, i, f, o\}$ respectively represent the $block~input$, ${input~gate}$, $forget~gate$, and $output~gate$; $\sigma$ is the logistic sigmoid activation function, $h$ is the hyperbolic tangent activation function, and $\odot$ denotes the point-wise product with the gate value. Moreover, the corresponding backward pass that required in the training stage can be found in the paper \cite{greff2015lstm}.

\subsubsection{Bidirectional LSTM recurrent structure}
\begin{figure}[!tp]
\centering
\includegraphics[width=70mm]{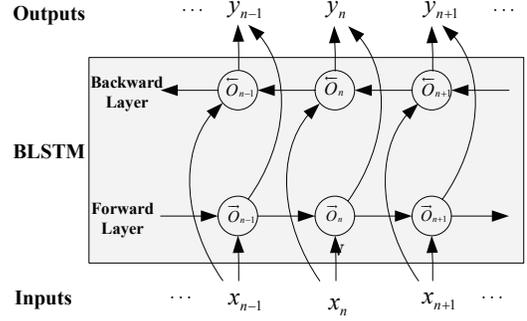}
\caption{Bidirectional LSTM recurrent structure diagram. }\label{fig5}
\end{figure}

The structure of LSTM memory block allows the network to store and retrieve information over long periods of time. For standard LSTM networks, there is an obvious drawback that they only have access to previous context but not to future context. This shortcoming can be overcomed by using bidirectional LSTM networks, which deal with the data in both directions with two separate hidden layers \cite{BLSTM-1}. These two hidden layers are connected to the same output layer. Thereby, the forward and backward contexts are learned independently from each hidden layer \cite{BLSTM-2}, as shown in Fig. \ref{fig5}. The bidirectional LSTM recurrent neural networks are implemented by the following functions:

\begin{equation}\label{equation 8}
\begin{split}
& {\overrightarrow O _n} = \Gamma ({W_{\overrightarrow I }}{x_n} + {W_{\overrightarrow O }}{O_{n - 1}} + {b_{\overrightarrow O }})\\
& {\overleftarrow O _n} = \Gamma ({W_{\overleftarrow I }}{x_n} + {W_{\overleftarrow O }}{O_{n + 1}} + {b_{\overleftarrow O }})\\
& {y_n} = {W_{\overrightarrow y }}{\overrightarrow O _n} + {W_{\overleftarrow y }}{\overleftarrow O _n} + {b_y}
\end{split}
\end{equation}
where ${\overrightarrow O _n}$ is the forward hidden sequence, ${\overleftarrow O _n}$ is the backward hidden sequence, and $\Gamma$ is implemented by Equ. \ref{equation 7}.

The proposed deep neural network basically has the ability to learn the long-range multi-aspect features by integrating the LSTM memory block and the bidirectional recurrent networks. To further construct a progressively higher level representations of multi-aspect features, we design a bidirectional LSTM network with three layers, whose size are $512$, $256$ and $128$, respectively. The completed forward pass process can be deduced from the above basic principles.

\subsection{Target Classification with Softmax}
After the multi-aspect feature learning, the outputs from the last bidirectional LSTM layer are normalized with the $softmax$ function:

\begin{equation}\label{equation 9}
P(c|{y_i}) = \frac{{\exp ({z_c}({y_i}))}}{{\sum\limits_{l \in [1,...,L]} {\exp ({z_l}({y_i}))} }}
\end{equation}
where ${y_i}$ is the output from the last hidden layer for corresponding target $c$, ${z_c}(\cdot)$ is the final activation function, and $l$ is the one of target classes.

Due to a multi-aspect image sequence may include several or tens of images, the decision over the whole sequence will reduce the recognition performance when the features of some sequence slices are not consistent with others'. Therefore, the $softmax$ layer with multiple $softmax$ function units is designed to solve this issue. Although the multi-aspect features are learnt from long-range contextual sequence slices, the final classification accuracy calculations are independent for avoiding the error spreading.

\section{Experimental Results}
\label{sec3}
For the proposed MA-BLSTM networks with multi-orientation spatial features, Matlab is used to implement the Gabor and TPLBP methods, MXNet library is used to reduce the feature dimensionality, and CURRENNT toolkit is employed to construct the bidirectional LSTM networks. MXNet is a multi-language deep neural networks library and runs on various heterogeneous systems, ranging from mobile devices to distributed GPU cluster \cite{mxnet}. CURRENNT is an open-source parallel implementation of deep RNNs supporting GPUs acceleration \cite{currennt}. As to hardware environment, a workstation with two Intel Xeon E5-2683 CPUs and one NVIDIA Geforce GTX1080 GPU is employed as the training and testing platform. In order to evaluate the ATR performance, two categories of experiments are designed, respectively, the Standard Operating Condition (SOC) and the Extended Operating Condition (EOC) \cite{soceoc}. The SOC experiment is defined as the set of testing conditions ``very near" training conditions. The EOC experiments are defined to individually measure SAR ATR extensibility across EOCs: configuration, target versions, depression, number of classes, squint, aspect, serial number (SN) and so on. In this section, the SOC experiment of 10-class targets and the EOC experiments with configuration variants are performed and discussed.

\subsection{Experimental Data and Setup}

\begin{figure}[!tp]
\centering
\includegraphics[width=85mm]{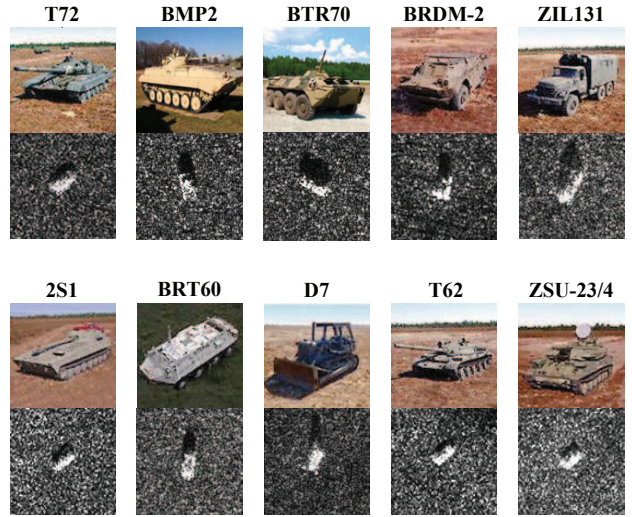}
\caption{Experiment samples of 10 targets: optical images (top) and SAR images (bottom). }\label{fig6}
\end{figure}

In SAR ATR community, the high resolution SAR data collected by the Defense Advanced Research Projects Agency (DARPA) and the Air Force Research Laboratory (AFRL) Moving and Stationary Target Acquisition and Recognition (MSTAR) program is almost the unique data set for developing and evaluating the ATR algorithms \cite{MSTARATR}. Hundreds of thousands of SAR images containing ground targets were collected, including different target types, aspect angles, depression angles, serial number, and articulation, and only a small subset of which are publicly available on the website \cite{DLSARJYQ,MSTAR}. The data set consists of X-band SAR images with $1$ foot by $1$ foot resolution and $0\sim360^\circ$ aspect coverage, which contains ten types of vehicle targets, as shown in Fig. \ref{fig6}.

For the SOC experiment, all the ten types of vehicle targets are used to evaluate the multi-classification performance. In order to meet the requirement of SOC experiment, namely the close situation between the training and testing sets, the image sets with same serial number and different depression angle are selected as the training and testing sets, which are respectively constructed for several image sequences, as listed in Tab. \ref{Tab1}. Each target has views at $15^\circ$ and $17^\circ$ depression angles. The data in depression $17^\circ$ are used for training and the other for testing. Except for the depression angle, the aspect angles of these two sets may be different. Otherwise, the employed training set is only hundreds level, which cannot afford the training overhead for very deep networks. But it is suitable for relative shallow networks like ten layers or less. In the EOC experiment with configuration variant, four targets in $17^\circ$ depression angle (T72, BMP2, BRDM2 and ZSU23/4) are selected as the training set, T72 targets with five different serial number and two different depression angles are selected as the testing set, as listed in Tab. \ref{Tab2}.

\begin{table}[!tp]
\caption{Training and Testing Images for the SOC Experimental Setup}
\label{Tab1}
\centering
\begin{tabular}{cccccc}
\hline
            &          &    Train &     &Test  &  \\
\hline
    Class    &  Serial No.  & Depression & Quantity  & Depression & Quantity  \\
\hline
T72 & 132 &  $17^\circ$ &232   & $15^\circ$  & 195 \\
BMP2 & 9563&  $17^\circ$ &232 & $15^\circ$ & 195   \\
BTR70 & C71 & $17^\circ$ &232 & $15^\circ$ & 195  \\
BRDM2 & E71 & $17^\circ$ &232& $15^\circ$  & 195 \\
ZIL131 & E12 & $17^\circ$ &232& $15^\circ$  & 195 \\
2S1 & B01 & $17^\circ$ &232& $15^\circ$  & 195 \\
BTR60 & 7532 & $17^\circ$ &232& $15^\circ$  & 195 \\
D7 & 3015& $17^\circ$ &232& $15^\circ$ & 195 \\
T62 & A51 & $17^\circ$ &232& $15^\circ$  & 195 \\
ZSU23/4 & D08 & $17^\circ$ &232& $15^\circ$  & 195 \\
\hline

\end{tabular}
\end{table}

%
    \renewcommand{\multirowsetup}{\centering}
    \begin{table}[!tp]
    \footnotesize
    \caption{Testing Images for the EOC Experimental Setup}
    \label{Tab2}
    \centering
        \begin{tabular}{cccc}
        \hline
        Class    &  Serial No.  & Depression  &  Quantity  \\
        \hline
        \multirow{5}{1cm}{T72} & S7 & $15^\circ$,$17^\circ$ &419 \\
        & A32&  $15^\circ$,$17^\circ$ &572 \\
        & A62 & $15^\circ$,$17^\circ$ &573\\
        & A63 & $15^\circ$,$17^\circ$ &573\\
         & A64 & $15^\circ$,$17^\circ$ &573\\
        \hline
        \end{tabular}
\end{table}

Next, we will simply describe the experiment configuration. The size of the input images is $128\times128$. The Gabor filter utilizes 6 directions, respectively $0$, $\pi/6$, $\pi/3$, $\pi/2$, $2\pi/3$ and $5\pi/6$. Thus, the size of each input image becomes $128\times128\times6$ after the Gabor filtering. For the TPLBP feature extraction, the radius $r$ is set to 12, the patch number $S$ is set to 8, the patch size $w$ is set to 3, $\alpha$ is 1, and the block size of histogram measuring is $20\times20$. The final extracted multi-orientation spatial feature dimension of each image is 75264. Correspondingly, the size of MLP input layer, hidden layer and output layer are 75264, 1024, 10 in SOC experiment, and 4 in EOC experiment. Finally, the sizes of three bidirectional LSTM layers are $512$, $256$ and $128$, respectively.

\subsection{Recognition Performance Under SOC}

\begin{table*}[!tbp]
\scriptsize

\caption{Confusion Matrix for the SOC experiment}
\label{Tab3}
\centering
\begin{tabular}{cccccccccccc}

\hline
   Class & T72    &  BMP2  & BTR70 & BRDM2  & ZIL131 & 2S1 & BTR60 &D7 & T62 & ZSU23/4 & Accuracy(\%)\\
\hline
T72  & 195 &  - &  - & -  & -  & - & - & - & - & -  & 100\\
BMP2 & - &  195 &  - & -  & -  & - & - & - & - & -   & 100 \\
BTR70 & - &  - & 195 & -  & -  & - & - & - & - & -    & 100\\
BRDM2 & - &  - &  - & 195  & -  & - & - & - & - & -      & 100\\
ZIL131 & - &  - & - & -  & 195  & - & - & - & - & -   & 100 \\
2S1   & - &  - &  - & \textbf{1}  & -  &194 & - & - & - & -  & 99.49 \\
BTR60 & - &  - &  - & -  & -  & - & 195& - & - & -  & 100\\
D7 & - &  - &  - & -  & -  & - & - & 195 & -& -  & 100 \\
T62 & - &  - &  - & -  & \textbf{1}  & - & - & - & 194 & -  & 99.49 \\
ZSU23/4 &  - &  - & -  & -  & - & - & - & - & -  & 195  & 100 \\
\hline
Total  &  &   &   &   &   &  & &  & &   & \textbf{99.90} \\
\hline
\end{tabular}
\end{table*}

In the recognition performance experiment under SOC, the 10-target classification is employed to evaluate the proposed multi-aspect-aware bidirectional LSTM (MA-BLSTM) networks. The attribute descriptions of training set and testing set are listed in Tab. \ref{Tab1}. Although each target image has $16384$ pixel intensity values, the size of input features for bidirectional LSTM is $1024$. According to our empirical study, the best learning rate is $10^{-7}$, and the recognition results will converge within thousands of training cycles. The confusion matrix of the SOC experiment is listed in Tab. \ref{Tab3}. The first column indicates the actual categories, and the middle ten columns represent the number of images classified to $10$ targets. It can be seen that the recognition performance of MA-BLSTM is excellent. There are only two images that are classified incorrectly. One $2S1$ image is confused as $BRDM2$ target, and one $T62$ image is confused as $ZIL131$ target. Except for these two targets, other targets are all recognized with $100\%$ accuracy.

\begin{figure}[h]
\centering
\includegraphics[width=90mm]{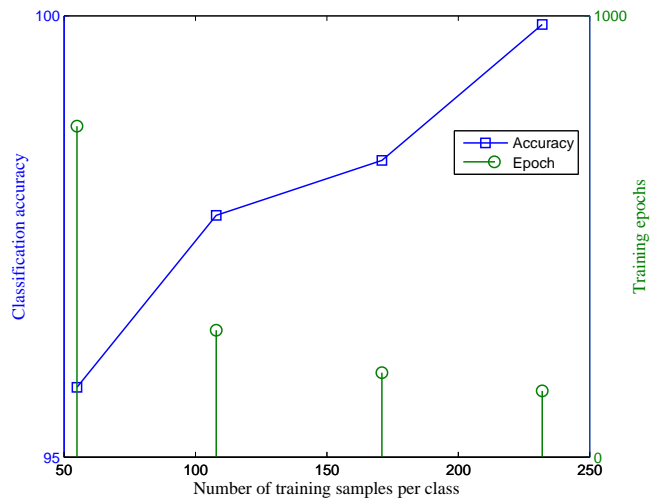}
\caption{Training and recognition performance with different numbers of training samples. }\label{fig8}
\end{figure}

Many of recognition and deep learning algorithms require significant training samples, thus it cannot be directly applied to practical applications when training data set is sparse. In current ATR domain, the rapid learning algorithms with limited training data are highly demanded. In order to evaluate the sensitivity of MA-BLSTM to the size of training samples, another experiment considering different numbers of training samples and same testing samples is carried out. The testing samples and baseline training samples are listed Tab. \ref{Tab1}, respectively, have $1950$ and $2320$ images in total. Three different percentages of base training samples are employed to investigate the small samples recognition capability. As shown in Tab. \ref{Tab7}, the classification accuracy only decreases $4.11\%$ when the number of training sample per class is changed from $232$ to $55$. The results show that MA-BLSTM has a good classification performance for the case of small samples, and is not very dependant on the size of training samples. On the other side, the training time cost increases by $5$ times with the decrease of training data, as shown in Fig. \ref{fig8}. Nevertheless, the training efficiency issue is not a big problem, and can be solved by introducing hardware acceleration like multiple GPUs or fast training strategy. In general, the multi-aspect feature learning collects more essential characteristics of SAR targets, and makes the MA-BLSTM more suitable for the small sample set recognition and practical applications.

\begin{table}[h]
\footnotesize
\caption{Classification Performance with Sparse Training Samples }
\label{Tab7}
\centering
\begin{tabular}{ccccc}

\hline
   Training percentage  & $100\%$ & $74\%$   &  $47\%$  & $24\%$ \\
\hline
   Training samples per class  & $232$ & $171$   &  $108$  & $55$ \\
   Training epochs  & $150$ & $191$   &  $287$  & $750$ \\
   Classification accuracy & $99.90\%$ & $98.36\%$   &  $97.74\%$  & $95.79\%$ \\
\hline

\end{tabular}
\end{table}

In practical applications, the number of multi-aspect images may be less than the experiments employed, namely full-aspect coverage. On the other hand, the researchers from Swedish Defence Research Agency (FOI) state that the variation in aspect angle during the illumination must be $180^\circ$ minimum to ensure that the space-varying signature is covered when the direction for the front and the rear of a vehicle is unknown \cite{CSAR-SWE}. Therefore, we want to find out the basic requirements for sequence length and aspect coverage. Supposing the MA-BLSTM is trained with all-aspect training data in advance, four experiments that respectively simulate the different SAR observation scenarios are designed to discuss the recognition performance with limited aspect images. Specifically, the differences with the 10-class experiment are only the length and number of test sample. As Tab. \ref{TabLOS} listed, the first two experiments represent the SAR observations with dense aspect interval and different coverage, the last two experiments indicate the multiple SAR observations with sparse aspect interval and full-aspect coverage. From the results, it can be seen that the aspect coverage is more important than the multi-aspect collection density. In other words, it is better to select dense observation strategy in narrow aspect coverage condition, and employ sparse observation strategy in the wide aspect coverage condition. So, supposing several UAV SAR systems observe one target over $180^\circ$ aspect range in practical situation, high classification accuracy still can be achieved with our trained MA-BLSTM framework.

\begin{table}[h]
\footnotesize
\caption{Classification Performance with Limited Aspect Angles}
\label{TabLOS}
\centering
\begin{tabular}{ccccc}

\hline
   Range & Interval      & Quantity &Sequence &accuracy\\
\hline
   $0-360^\circ$ &  $6-9^\circ$  & 40-60     &1  & $99.90\%$ \\
   $0-180^\circ$ &  $6-9^\circ$     &20-30   &1  & $99.90\%$ \\
   $0-360^\circ$ &  $60-90^\circ$     &4-6   &1  & $99.90\%$ \\
   $0-180^\circ$ &  $30-45^\circ$     &4-6   &1  &  $97.30\%$ \\

   \hline

\end{tabular}
\end{table}

\begin{figure}[!bp]
\centering
\includegraphics[width=85mm]{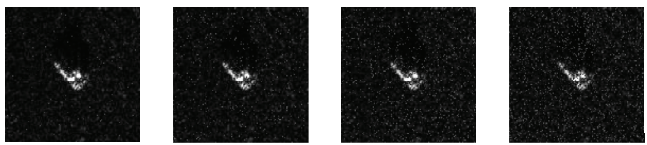}
\caption{Illustration of noise-contaminated images with noise levels of 1\%, 5\%, 10\% and 15\%. }\label{fig7}
\end{figure}

Due to the Gabor and TPLBP features are employed in MA-BLSTM as the single aspect slice of multi-aspect features, the expected antinoise performance may be solid. In \cite{DLSARJYQ}, the noise-contaminated experiment is performed to evaluate its performance. Similarly, we randomly choose pixels from the testing images by a certain percentage, denoting the noise level. Then we replace their intensity values with pseudo-random numbers subjected to a uniform distribution, as shown in Fig. \ref{fig7}. With the same trained MA-BLSTM by 10-target data set, the recognition accuracy of four different noise levels are listed in Tab. \ref{Tab4}. Although the recognition accuracies of two networks on 10-target data set are close, the antinoise performance of MA-BLSTM is greatly improved. When the noise level increases $15\%$, the accuracy decreases about $5.5\%$ in MA-BLSTM, but $45\%$ in the all-convolutional neural networks (A-ConvNets) \cite{DLSARJYQ}.

\begin{table}[h]
\footnotesize
\caption{Comparison Of Antinoise Performance  }
\label{Tab4}
\centering
\begin{tabular}{cccccc}

\hline
   Noise & $0$ & $1\%$   &  $5\%$  & $10\%$ & $15\%$ \\
\hline
A-ConvNets \cite{DLSARJYQ} & 0.9913 &  0.9176 &  0.8852 & 0.7584  & 0.5468 \\
\hline
MA-BLSTM  & 0.9990 &  0.9948 &  0.9810 & 0.9774  & 0.9441 \\
\hline

\end{tabular}
\end{table}

\subsection{Recognition Performance Under EOC}
It is meaningful and comparable that experiments should be done with conscious isolation of EOCs, and such sensitivity results can be extended to operational scenarios \cite{MSTARATR}. In the EOC experiment, the isolated factor is target configuration variant, which is also adopted by other ATR methods for testing. The confusion matrix for EOC experiment is listed in Tab. \ref{Tab5}. As for the recognition performance, there are only $11$ images misclassified of all the $2170$ testing samples, and $99.59\%$ classification accuracy is achieved. The reason for the confusion of BMP2 and T72 may be the similar appearance and image representation. To sum up, the EOC experimental results indicate two meanings: one is that the proposed MA-BLSTM still keeps a high recognition performance in configuration variant scenario, and the other is that the T72 is more confused with BMP2, which is accorded with the result of paper \cite{DLSARJYQ}.

\renewcommand{\multirowsetup}{\centering}
\begin{table}[h]
\footnotesize
\caption{Confusion Matrix for the EOC Experiment}
\label{Tab5}
\centering
\begin{tabular}{p{0.6cm}p{1.2cm}p{0.6cm}p{0.6cm}p{0.8cm}p{0.6cm}p{1.2cm}}
\hline
   Class & Serial No.   & BMP2  & BTR70 & BRDM2 &  T72 & Accuracy(\%)\\
\hline
 \multirow{5}{0.6cm}{T72}
        & S7 & 7& -&  - &412& 98.33 \\
        & A32 & 1& -&  - &571& 99.83 \\
        & A62 & 3& -&  - &570& 99.48 \\
        & A63  & -& -&  - &573& 100 \\
         & A64  &-& -&  - &573& 100 \\
\hline
Total  &  &   &   &   &   &    99.59 \\
\hline
\end{tabular}
\end{table}

Furthermore, we also make a recognition performance comparison between the proposed MA-BLSTM and other typical ATR methods, which consist of SVM, Adaptive Boosting (AdaBoost) and A-ConvNets. Due to the employed training and testing samples are totally the same, the results of three typical ATR methods are cited from the corresponding papers. The average classification accuracies of four methods are listed in Tab. \ref{Tab6}. It can be seen that the deep learning methods outperform the classical machine learning methods in two scenarios. As for the two deep learning methods, the proposed MA-BLSTM is slightly superior to the A-ConvNets because of the multi-aspect space-varying scattering feature learning, which is essentially the key difference between SAR and optical image recognition.

\begin{table}[!htbp]
\footnotesize
\caption{Classification Accuracy Comparison of MA-BLSTM and Other Methods}
\label{Tab6}
\centering
\begin{tabular}{ccc}
\hline
    Method    &  SOC(\%)  & EOC(\%)  \\
\hline
SVM \cite{JYQsar31} & 90 &75   \\
AdaBoost \cite{JYQsar31}& 92 &78   \\
A-ConvNets \cite{DLSARJYQ}& 99.13 &98.93   \\
\hline
MA-BLSTM & \textbf{99.9}0 & \textbf{99.59}  \\
\hline
\end{tabular}
\end{table}

\section{Conclusions}
\label{sec4}
In this paper, a multi-aspect-aware bidirectional LSTM networks based SAR ATR framework has been proposed. The Gabor and TPLBP methods are employed to implement the initial static feature extraction considering multi-orientation spatial information. After that, a fully-connected MLP neural network is utilized to reduce the feature dimensionality. To learn the broad contextual feature simultaneously when mapping between input and output sequences, the bidirectional LSTM architecture is introduced to exploit the space-varying scattering feature among different aspects and ensure a sufficient range of context coverage. By benefiting from the multi-aspect information learning, the proposed MA-BLSTM method can further improve the performance of existing machine learning and deep learning methods. Specifically, it can not only reach the recognition accuracy of $99.9\%$ in 10-target classification issue, but also show the good antinoise and anti-confusion performance. Furthermore, the input of existing deep neural network based SAR ATR methods is usually one image, which is similar with the optical image recognition. In order to learn the special space-varying scattering characteristic of SAR target, the multi-aspect image sequence is taken as the input sample of MA-BLSTM, which is a essential design for SAR ATR problem. It can be expected that the bidirectional LSTM neural network that is capable of learning the multi-dimensional features of SAR target may play a key role in enabling better discrimination of targets in a confusing environment.

\section{Acknowledgements}
This work was supported by the National Natural Science Foundation of China under Grant No.~$61501018$, Grant No.~$61571033$, Grant No.~$61431018$, by the Beijing Natural Science Foundation under Grant No.~$4164093$, and by the National Major Research High Performance Computing Program of China under Grant No.~2016YFB0200300.

\normalsize
\bibliographystyle{ieeetr}
\bibliography{raw-data-paper}

\end{document}